\documentclass[letterpaper, 10 pt, conference]{ieeeconf}  

\IEEEoverridecommandlockouts                              
 
\usepackage{graphicx} 
\usepackage{epsfig} 
\usepackage{amsmath} 
\graphicspath{{./Figures/}}
\usepackage{gensymb}
\usepackage{float}
\usepackage{subfig}
\usepackage{breqn}
\usepackage{color}
\usepackage{wrapfig}
\usepackage[hidelinks]{hyperref}

\title{\bf Estimating Human Intent for Physical Human-Robot Co-Manipulation }
 
\author{Eric C. Townsend \: Erich A. Mielke \: David Wingate \: Marc D. Killpack  
}
 
\begin{document}
 
\maketitle
\thispagestyle{empty}
\pagestyle{empty}
 
\begin{abstract}
Human teams can be exceptionally efficient at adapting and
collaborating during manipulation tasks using shared mental
models. However, the same shared mental models that can be used by
humans to perform robust low-level force and motion control during
collaborative manipulation tasks are non-existent for robots. For
robots to perform collaborative tasks with people naturally and
efficiently, understanding and predicting human intent is
necessary. However, humans are difficult to predict and model. We have
completed an exploratory study recording motion and force for 20 human
dyads moving an object in tandem in order to better understand how
they move and how their movement can be predicted. In this paper, we
show how past motion data can be used to predict human intent. In
order to predict human intent, which we equate with the human team's
velocity for a short time horizon, we used a neural network. Using the
previous 150 time steps at a rate of 200 Hz, human intent can be
predicted for the next 50 time steps with a mean squared error of 0.02
$(m/s)^2$. We also show that human intent can be estimated in a human-robot dyad. This work is an important first step in enabling future
work of integrating human intent estimation on a robot controller
to execute a short-term collaborative trajectory.
 
\end{abstract}
 
\section{Introduction}
While robots have long been used in manufacturing, they are
increasingly gaining the capability to work in unstructured and
dynamic environments. In the future, this will include applications
related to logistics, healthcare, agriculture, disaster response, and
others. However, robots will also be required to successfully interact
more naturally with human teammates. Interacting with people in a way
that is helpful and intuitive requires that the robots both act
predictably and be able to predict human intent. Specifically, in this
paper we aim to create a model for predicting human intent in a
collaborative object carrying task that a robot could use to be a
more responsive and intuitive teammate.

\begin{figure}[hbt]
  \centering
  \includegraphics[width=.9\linewidth]{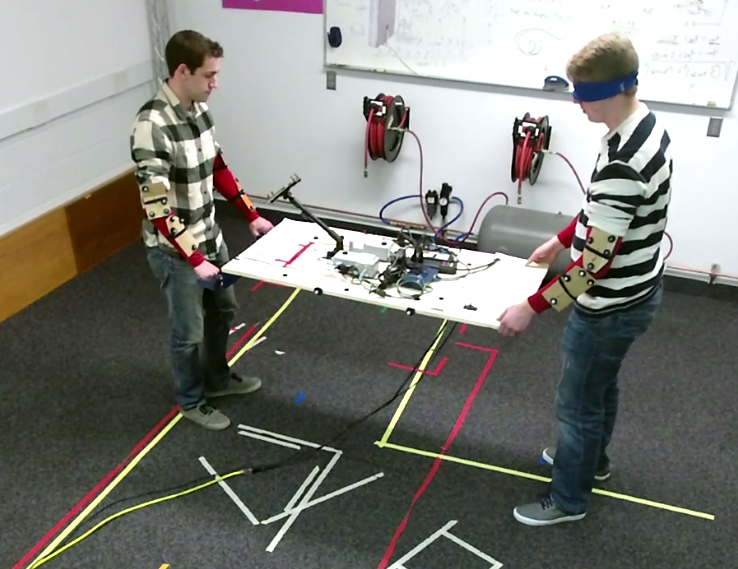}
  \caption{A leader and a blindfolded follower performing a table carrying task.}
  \label{fig:dyad}
\end{figure}

In this paper, we call physical human-robot interaction for the
purpose of collaboratively manipulating an object co-manipulation. In
order to understand human-human co-manipulation, we ran an exploratory
study with 20 human dyads. Each dyad moved a long board representing a
table as we measured their movement and forces on the board as in Figure \ref{fig:dyad}. Many
previous studies have been done on human movement in which one or two
people move in tandem. Many of these are done in haptic simulations or
with limited degrees of freedom in order to isolate specific
behaviors. These studies have given significant insight on things like
minimum jerk motion, negotiation of roles, and task-specific
movements. However, we expect that due to the nature of this past work,
which has mostly examined a limited number of degrees of freedom,
there are limitations to how those results can be extrapolated for
general purpose six dimensional co-manipulation tasks. We therefore
assert it is also necessary to perform studies with natural and
realistic human movement without limiting degrees of freedom. This would validate what has been learned in other studies and to allow further
insight and direction for human-robot co-manipulation controller
development.

There are many sources of information that a robot could use to
predict human intent, including motion, force, partner posture, and verbal
communication among others. In our study, we chose to focus on motion
and force. We hypothesized that these variables were the most
fundamental and easiest to interpret for a robot in order to predict
what the person intends to do. In another paper \cite{Mielke2017}, we
discuss insights gained from the force information in this study and
how force data can predict initiation of future motion. While in this
paper we use past motion data only to predict where the co-manipulated
object will move next. The interaction between force and motion in a
co-manipulation task is complicated and important. In this preliminary
work, we have attempted to isolate and explore each physical quantity
before we later expect to integrate insights and models from both for
future robot controller development.
 
Many models have been created to understand how people move and
collaborate. In this work, we choose to train a neural network because
they can be used to approximate any function and have been used to
forecast time series \cite{Frank2001}. Therefore, as long as the
future human motion is in fact predictable based on the input
variables, it should be possible for the neural network to predict the
human motion.

After training the neural network, we show that the neural
network is also able to predict future intent of a human-robot
dyad. We compare the prediction of the neural net to the actual motion
of the robot. It is impossible to decouple the intent of the two
people in a dyad. Therefore, in order to actually use our intention
estimation for future controllers, it is important to show that the
neural network can still predict the motion when one of the humans is
replaced with a robot. In this paper, we do show that with a simple
impedance controller on the robot, the neural network is able to
accurately predict what the human-robot team will do.

The contributions of this paper include the following:
\begin{itemize}
\item Determination that past motion of a human dyad is all that is
  necessary to predict human intent for at least a limited time
  horizon, where we define ``intent'' as how the team will move next.
\item Development of a neural network to predict human intent based on
  past motion of the human dyad.
\item Validation of the human intent prediction in a human-robot dyad.
\end{itemize}
 
The paper is organized as follows. Section \ref{sec:related-work}
describes related work on human-robot interaction and intent
modeling. The study on human-human interaction is described in Section
\ref{sec:exploratory-study}. The structure and training of the neural
network is describe in \ref{sec:neural-network}. In Section
\ref{sec:robot-description} we describe our robot platform and the
validation of the estimate of human intent on our robot. Finally, we
discuss the results in Section \ref{sec:results-and-discussion}.
 
\section{Related Work}
\label{sec:related-work}
Many different control methods have been used in the last 20 years for
human-robot cooperation. The first research in this area developed
from what had been done with exoskeletons, because it involved both a
robot and a human working together to manipulate an object.  One of
the first controllers for cooperative manipulation of an object by
robots and humans was an impedance controller that could be used with
any number of robots and humans holding the object
\cite{Kosuge1993}. In general, the advantage of human-robot
collaboration is that humans provide intelligence and dexterity while
robots may provide strength and stability \cite{Kazerooni1988}.

To improve human-robot interaction, human-human interaction has been
studied to in order to understand how people communicate
haptically. Ikeura et al. showed that when two people move an object,
but only one knows the task, the applied force is highly correlated to
the velocity and therefore can be modeled by a damping element. Of
importance was the result that the spring and mass terms were much
less important than the damping terms \cite{Ikeura1994}. Ikeura later
showed that a constant damping term does not work well for changing
velocities as it slows the human-robot pair significantly compared to
the human-human pair. Variable damping that depends on velocity was
proposed to have both fast motion and accurate positioning when using
an impedance model \cite{Ikeura1995, Ikeura1997}. One issue with this
model is that it was determined and validated using a one-dimensional
trajectory at a short distance. This makes it difficult to generalize
for cooperative motion during a general six degree of freedom task.
Additionally, other work has shown that a damping controller is not
clearly superior to other methods. One of the proposed measures used
to determine the effectiveness of a human-robot interaction controller
was the interaction force between the human and the robot. As this
force is usually counted as wasted energy, the hypothesis was that
humans naturally try to minimize it.  According to Ito et al., a
controller for human-robot cooperation should minimize this
interaction force as well \cite{Ito1999}. However, damping controllers
tend to have high internal forces. Motion is also not tracked as
accurately because the leader essentially drags the robot into place
instead of being assisted by the robot.

Many insights into human-robot interaction have been gained from the
field of neuromechanics. Flash and Hogan showed that the human arm
tends to follow a minimum jerk trajectory in many situations, which
means humans try to move in a smooth manner. This is accurate
especially when force is low and high speed is not an objective
\cite{Flash1985}. Other researchers showed how humans adapt to unknown
or unstable dynamics and variable impedance controllers have been made
based on this principle as well \cite{Yang2011, Kadiallah2012,
  Ganesh2010}. Ikeura used the minimum jerk model to calculate a
damping constant \cite{Ikeura1998} while Maeda et al. used the minimum
jerk model directly to predict what the human is trying to do. Their
controller estimates the final time and position that the human is
attempting to reach based on the past few steps and uses position
control to follow that trajectory along with impedance control to
adapt to errors \cite{Maeda2001}. Kheddar et al. have found that
minimum jerk does not apply to all cooperative movements. One example
is when motions are longer and include walking and not just arm
movement, there are phases of constant velocity along with phases of
constant acceleration. They used this knowledge to create a controller
based on phases of constant velocity that are only changed when a
certain force threshold is surpassed \cite{Miossec2008, Bussy2012}.

The minimum jerk model is one example of an invariant model, where by
finding a model for a task, the dynamics can be simplified. Many models
for a controller for cooperative manipulation only have a few changing
parameters that include initial and final time and position. They may
also include a maximum velocity or other parameter. With just a few
parameters, the entire trajectory can be described. One approach to
determining these models has been programming by demonstration
\cite{Evrard2009a}.

Finally, past work has shown that the robot should be predictable. The
human should not be surprised by what the robot does. Aside from being
predictable, if the robot takes any leadership in the task, such as
keeping an object from rotating, it should do so in a way that is
understandable to the human \cite{Dragan}. Chipalkatty et al. showed
that there is an advantage to making a controller simple enough that a
human can learn it quickly. If it is complex and constantly adapting,
there is a risk the human will become confused and the cooperation
will be inefficient \cite{Chipalkatty}.

The work that has been described so far is representative of the
current research in the field of physical human-robot interaction for
co-manipulation. We propose that most good human-robot interaction
controllers for cooperative manipulation of an object can be judged on
certain factors including low interaction forces, quick and accurate
motion tracking, safety, simplicity, robot initiative, and
predictability. However, very little of the past work has been based
on real performance data from human-human co-manipulation trials for
unconstrained tasks. The purpose of this paper was to use real
human-human data and attempt to develop an estimator that allows us to
predict human motion intent for a given set of tasks.
 
\section{Exploratory Study}
\label{sec:exploratory-study}
 
As described in more detail in \cite{Mielke2017}, we had 20 human
dyads perform a series of table carrying tasks. Figure \ref{fig:dyad} shows one task being performed and
video of the most complex task can be seen at
\url{https://youtu.be/DAbLRDN20yE}. There were 12 different tasks that
were each performed 3 times. Our purpose in doing this study was to
understand how people collaborate moving an object with no constrained
degrees of freedom. We wanted to verify that the principles other
researchers have shown in past studies with simpler tasks generalize
to all co-manipulation tasks. Additionally, we expected to identify
new principles from human-human data that could improve future control
development. The tasks chosen were based on this idea, and the tasks
represented several different types of common motion. Many of the
tasks were designed to isolate certain behavior (such as while
translating or rotating an object) while others were open-ended. The
tasks included lateral translation, forward/backward translation,
rotation, ``hallway'' navigation, lifting over obstacles, etc. These
were chosen to represent the variety of tasks that people would do in
real life.

During the experiment, each partner was assigned a role as leader or
follower. The leader was then given instructions on where to move the
table. In half of the tasks, the follower was blindfolded and talking
was prohibited. This can be thought of as a baseline for any robot
controller that does not use vision-based feedback. In the other half
of the trials, talking was permitted and there was no blindfold. The
performance in these trials could be considered the long-term
objective for performance and something against which we can benchmark
our future human-robot controllers.

The table and each participant were set up with motion tracking
markers. There were two force/torque sensors between the table and the
handles used by the leader. A Microsoft Kinect also tracked the pose
of the participants. Data was recorded from motion capture and the
force/torque sensors at 200 Hz. The Kinect recorded data at 15 Hz.

A board was used as a table-like object to be transported by the
dyad. The table had two handles on one side of the board. Each handle
was connected to the board with a force/torque sensor in series. The
table also had motion capture markers in order to get ground truth
data about the pose of the table at all times. The table also carried
an tablet that was oriented so that only the leader could see it. This
tablet was controlled by the experimenters to show the leader the
current task. It showed the leader where they were starting and where
they would need to go in the task. The tablet had images with markings
that corresponded to colored markings on the floor that indicated
where to move.

Finally, the table also had a power strip that powered the
force/torque sensors and the tablet. The power strip and ethernet
cables were each connected to the wall off-board the table. One of the
researchers was tasked with ensuring that these did not get in the way
of the participants and that there was sufficient slack so that no or
little additional force was placed on the table.
 
\section{Neural Network}
\label{sec:neural-network}
As a first approach to developing a nonlinear estimator of human
intention, we formulated a neural network using the Google Tensorflow
API. 

\subsection{Variables}
Originally, we expected to train the neural net on the force and
motion data from the exploratory study. As we trained the neural, we
discovered that force data was difficult to use and gives inferior
prediction of human intent. 
There are two proposed reasons for the low quality
prediction: 1) the force data was noisy, 2)the relationship between
the forces applied to the table by each person and the future motion
of the table is complex. The second point is in part because each
human dyad learns to communicate by force in their own way as shown in
\cite{Reed2006}.

We discovered that more accurate predictions could be made without
force. While longer term predictions may be aided with force
information, and initiation of motion as well, the motion data alone
is sufficient for an accurate short term prediction. This seems to
indicate that the follower acts or at least could be modeled as an
impedance while simply changing their equilibrium or set point during
co-manipulation. This is advantageous as it means that the estimation
of human intent can be used on robots without force sensors in the
wrist. In initial tests, we have only included translation from all of
our trials in the prediction and not rotation.

\subsection{Topology}
Our final neural network consists of 3 hidden layers each with 100
nodes. The process of choosing a neural network structure and other
parameters was not exhaustive and it is possible that better structure
and parameters could be obtained. However, we were limited in the
process for choosing neural net parameters as it can take on the order
of several minutes to train the neural net even when only training to
predict a single step. In the end, a simple structure using only
motion data as input worked well. Better structures, including methods
other than neural networks, may exist but our purpose in this paper is
to show that human intent estimation is possible. It was shown by
Chipalkatty et al. that more complex predictions of future movement
can actually decrease performance if they do not agree with what the human is trying to do. This is because predictions can be wrong and unpredictable. They found that it was more important that the
human understand what the robot will do next. \cite{Chipalkatty}. It is impoortant that the prediction be not only accurate, but also reliable. The
inputs to the neural network are 150 past steps of velocity and
acceleration in the x, y, and z direction, $\{x_{t-149},
x_{t-148}...,x_{t-1},x_{t}\}$. The outputs are the predicted velocity
and acceleration in the x, y, and z direction for 1 time step into the
future, $\hat{x}_{t+1}$, where $\hat{x}$ indicates a predicted value.

Our neural net formulation uses what Engel et al. describe as iterated
prediction \cite{Engel2004}. The neural network itself only predicts 1
time step into the future. Then, the prediction, $\hat{x}_{t+1}$, is
appended to the input to give $\{x_{t-149},
x_{t-148}...,x_{t-1},x_{t}\}$. The first step of the input is dropped
to obtain a new input of past motions for the neural net,
$\{x_{t-148}, x_{t-147}...,x_{t},\hat{x}_{t+1}\}$. The new data is
input into the neural net which outputs a prediction 1 step forward,
but 2 total steps into the future, $\hat{x}_{t+2}$. This is then
appended to the input. The process is repeated 50 times to obtain a
prediction of 50 steps, $\{\hat{x}_{t+1},
\hat{x}_{t+2}...,\hat{x}_{t+49},\hat{x}_{t+50}\}$. Because the outputs
of each prediction step become the inputs for the next, the inputs and
outputs must be the same variables.

\subsection{Training}
We pre-processed the data for the neural net to improve the
results. The velocity and acceleration data were scaled to have 0 mean
and standard deviation of 1 over the entire set of data. This was then
inverted on the output to show the results in their proper units. This
same scaling can be used on new data even though the mean and standard
deviation will be different. All trials with bad data were thrown
out. Where data was considered bad because of missing poses from
motion capture where too many motion capture markers were occluded
from the cameras. Each dyad was assigned randomly
to training and validation sets. 75\% of the data were assigned to the
training set and the other 25\% to the validation set. 

The neural net has to be trained in a special way in order to make the
iterated prediction $\hat{x}_{t+1}$ stable beyond the first step. This
process, described here, comes from \cite{Engel2004}. Batches of data
were created that randomly pulled in 32 sets of 150 steps of data from
the entire training set. Another 32 sets of 150 steps were
created from the validation set. The entire set of data consists of 2.5 million steps. The neural net was trained on new
training batches until the cost function reached a value less than a
threshold that we chose for 5 consecutive validation batches. We used
the mean squared error (MSE) for the cost function.

\begin{equation} \label{eqn:definition_Ix}
MSE = \sum_{n=1}^{32}(\hat{x}_{n,t+1}-x_{n,t+1})^2 
\end{equation}

The set in the batch is represented by $n$. Once the MSE was below the
threshold for 5 consecutive batches, a prediction was calculated for
every possible set of 150 steps of data. A new data set was created
that used 149 steps of real data and the prediction appended to the
end, $\{x_{t-148}, x_{t-147}...,x_{t-1}, x_{t}, \hat{x}_{t+1}\}$. The
neural network was then trained on a combined data set that included
the original data and the new data set that included the
prediction. Once this training was complete, a new data set was
created with 148 steps of real data and two predictions after it,
$\{x_{t-147}, x_{t-146}...,x_{t-1}, x_{t},\hat{x}_{t+1},
\hat{x}_{t+2}\}$. The same neural net was then trained again. This is
continued until the neural net will no longer converge in a reasonable
time. By training the neural network on data that includes
predictions, the stability of the prediction is improved.

The length of the prediction is limited by our computational resources
as we train the neural network. Our neural net predicts for 50 steps,
or .25 seconds into the future, because it was difficult for the
neural net to predict beyond that. We do not know if this is because
of a fundamental limit on the ability to predict after that amount of
time. We imagine there would be such a limit as the dyad has time to
make decisions on where to move, but determining how long it would be
for different teams is for future work. It could also be due to
limitations in the implementation of our neural network which could be
improved in the future.
 
The neural net was trained several times with the data split up randomly by dyad each time. This ensured that the neural net would be generalizable to an entirely new dyad.

\section{Robot Platform and Controller}
\label{sec:robot-description}
\begin{figure}[tb]
  \centering
  \includegraphics[width=.75\linewidth]{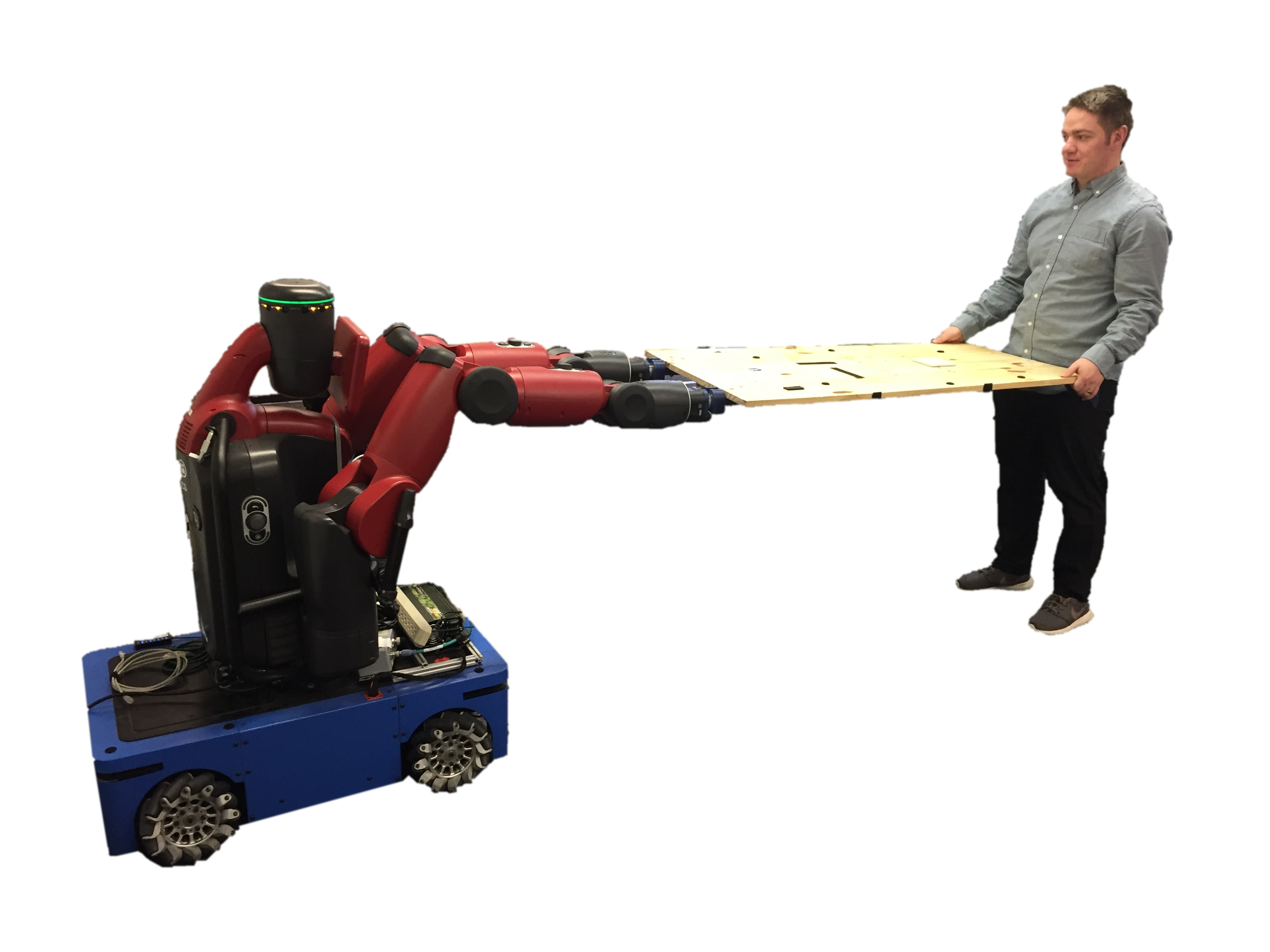}
  \caption{Rethink Robotics Baxter robot mounted on HStar Technologies AMP-1 holonomic base carrying the table with a person.}
  \label{fig:Megazord}
\end{figure}

The purpose of predicting human intent is to have the robot use the
prediction. Our robot platform for this research is a Rethink Robotics
Baxter robot mounted on an AMP-1 holonomic base from HStar
Technologies as seen in Figure \ref{fig:Megazord}. We chose to use a
holonomic base with mecanum wheels instead of something like a bipedal
robot in order to validate the human intent prediction at speeds
similar to two humans moving an object in every day life. This is
important to ensure that it works in real world applications as
limiting speed may affect the dynamics of the interaction.
 
We validated the neural network by predicting human intent as a human
and robot carry the table together. To perform this validation, we
implemented a controller that does not yet use the prediction of the
neural network. The Baxter arms are rigidly attached to the
table. Each arm is running a low impedance controller with a commanded
joint angle specified by the position of the arm relative to the table
before the task begins. When the arms are displaced, the mobile base
displaces by the same amount in order to put the arms in their
original pose relative to the base.

\section{Results and Discussion}
\label{sec:results-and-discussion}
\begin{figure}[H]
  \centering
  \includegraphics[width=0.9\linewidth]{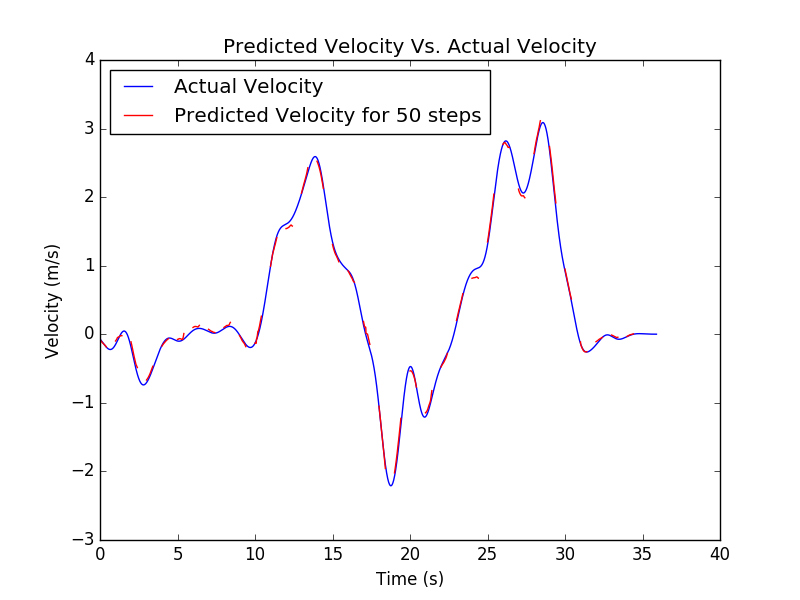}
  \includegraphics[width=0.9\linewidth]{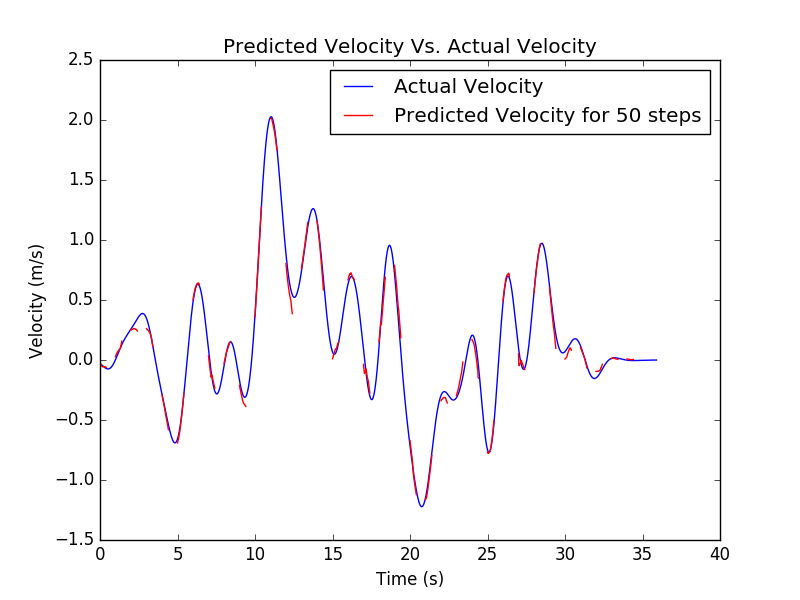}
  \includegraphics[width=0.9\linewidth]{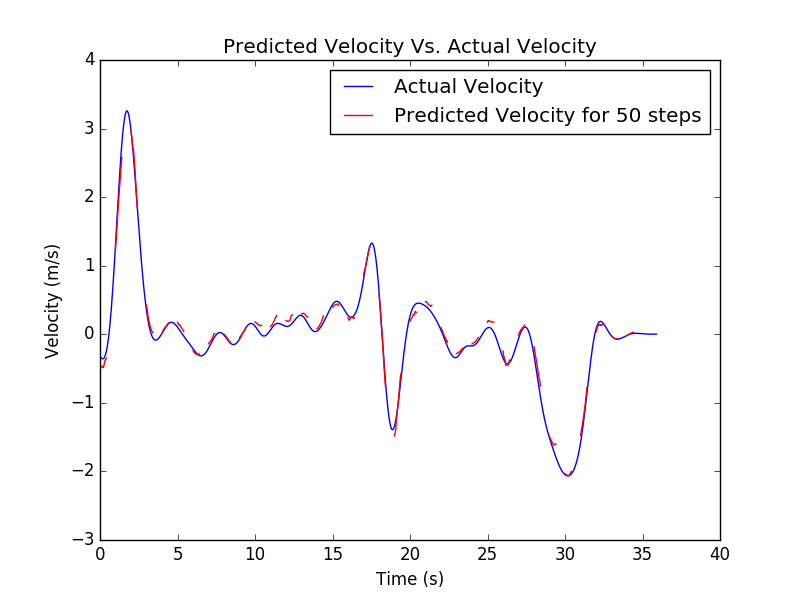}
  \caption{Comparison of velocity prediction to actual future data in the x,y, and z directions while a human dyad moves the table. Each red line is a separate 50 step prediction using the 150 steps before it. }
  \label{fig:prediction}
\end{figure}

\subsection{Neural Net Performance}
Figure \ref{fig:prediction} shows the neural network predictions of velocity in the x, y, and, z direction for a single representative task. The actual velocity is shown for the whole task in blue. The predicted velocity is shown in red starting every second and each one continues for 50 time steps or .25 seconds. As seen, the predictions are very accurate for that time scale. Here we only show velocity, but the acceleration data must also be predicted because each velocity prediction depends on the prediction of acceleration for the time step before it. Without acceleration data, the neural net performance degrades.

\subsection{Comparison to Polynomial Fit Predictor}
\begin{figure}[tb]
  \centering
  \includegraphics[width=1\linewidth]{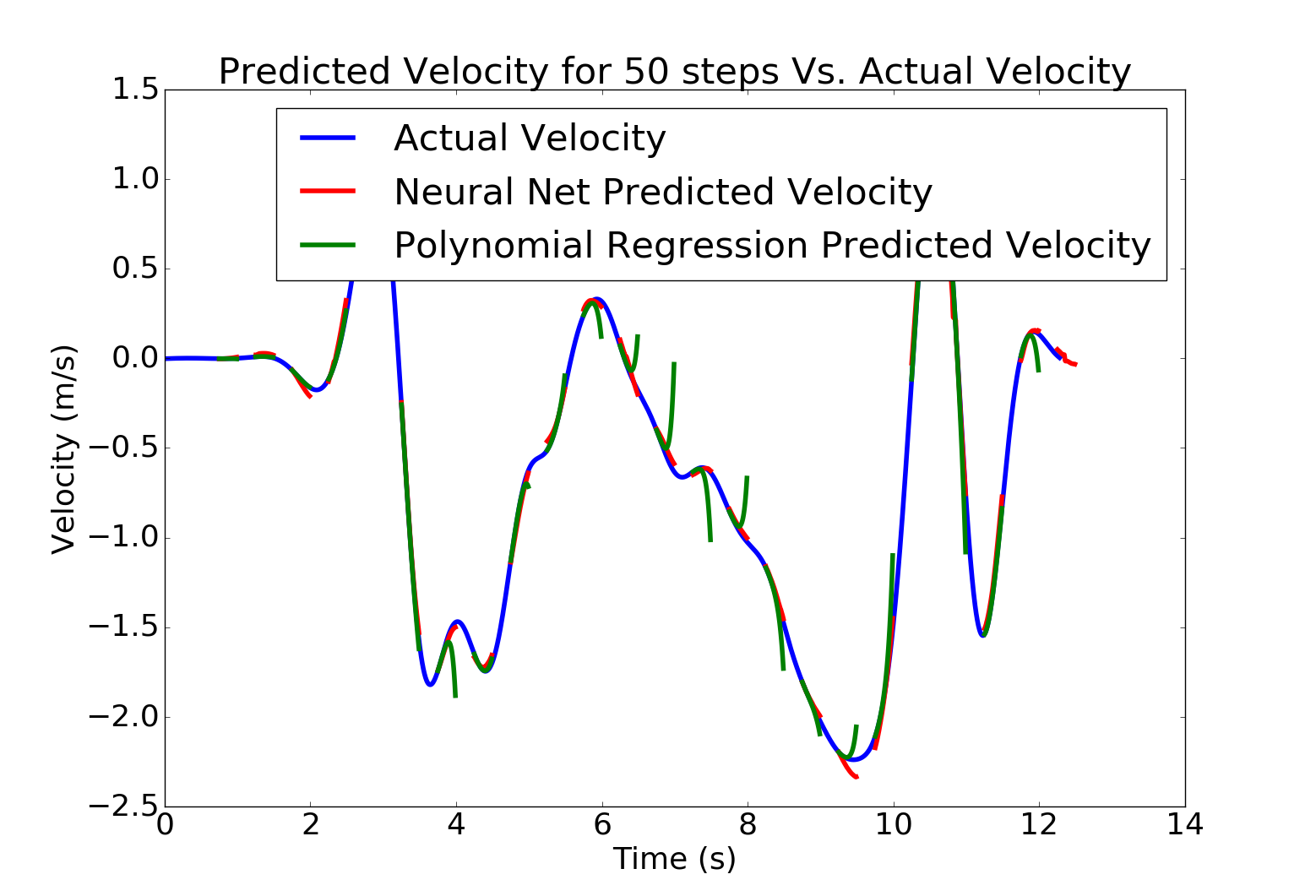}
  \caption{Predictions using the neural net and a polynomial estimator for velocity in one task. While both are accurate in most cases, in several cases the polynomial prediction is far from the actual data.}
  \label{fig:singletask}
\end{figure}

\begin{figure}[tb]
  \centering
  \includegraphics[width=1\linewidth]{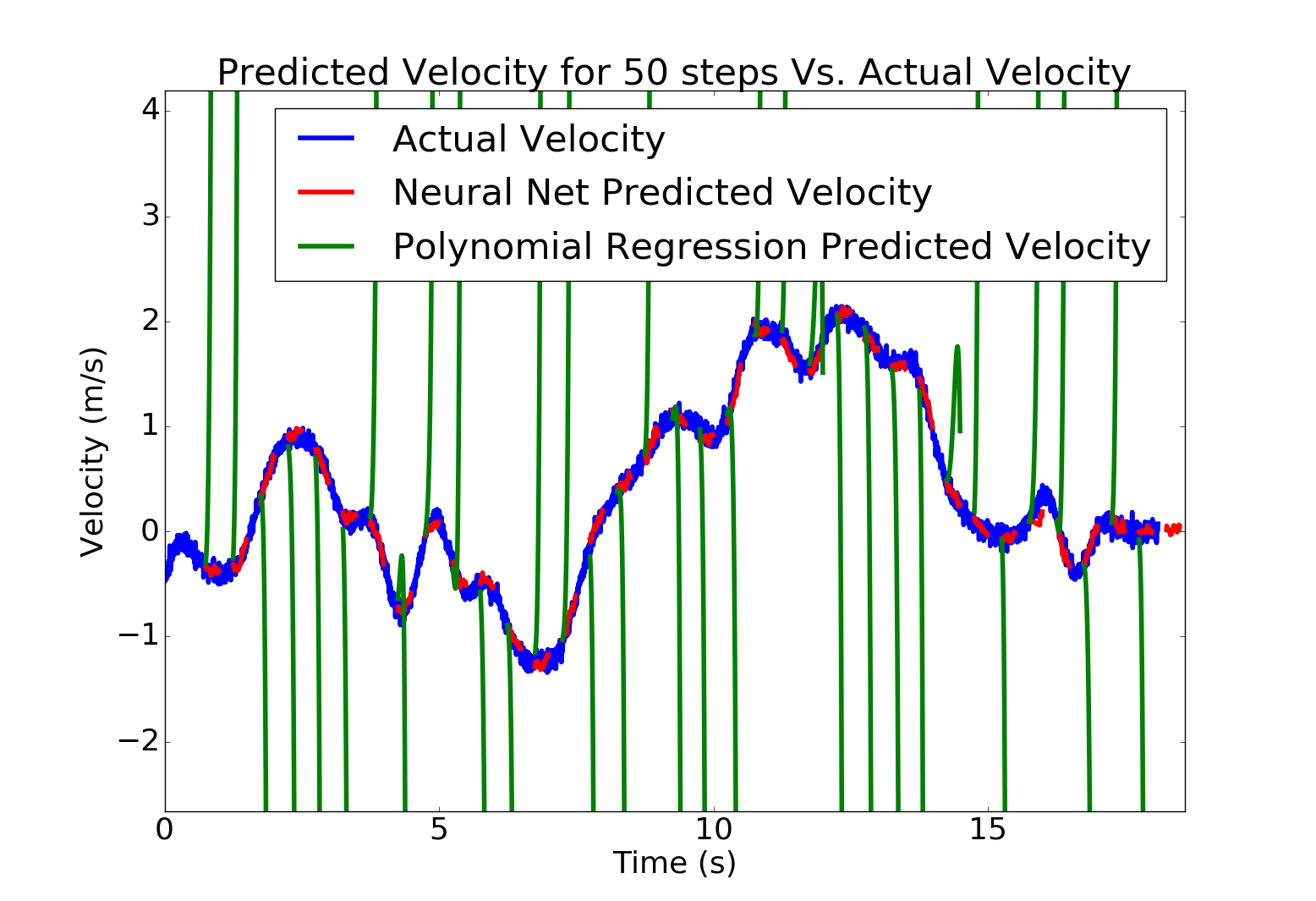}
  \caption{Predictions using the neural net and a polynomial estimator for velocity in one task where white noise has been added. The noise causes the polynomial prediction to go completely unstable while the neural network prediction is fairly robust.}
  \label{fig:singletasknoise}
\end{figure}

We developed a polynomial fit estimator to compare to our neural network. It also takes in the previous 150 steps and fits an 8th order polynomial. The 8th order was chosen because it had the best performance. The polynomial is then extrapolated forward 50 steps just like the neural net prediction. These are compared for a single random task in Figure \ref{fig:singletask}. The polynomial fit is accurate in many cases but is more prone to large errors. Figure \ref{fig:singletasknoise} shows how added noise affects the polynomial fit more than the neural network. This is a major issue as filtering the function to be smooth takes away time from the prediction so that it cannot estimate as far into the future.

\subsection{MSE of each Predictor}
\begin{figure}[tb]
  \centering
  \includegraphics[width=1\linewidth]{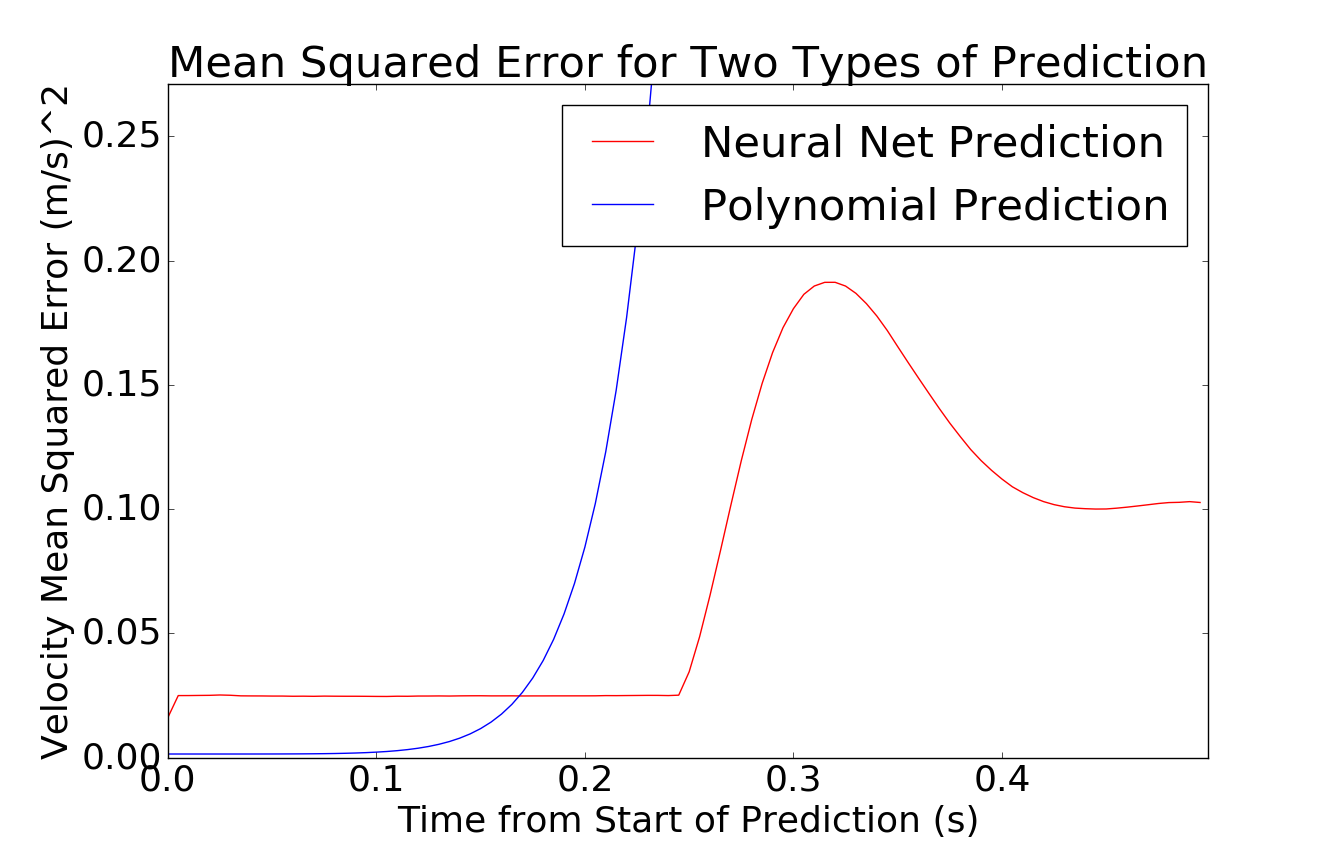}
  \caption{Mean squared error of the neural network prediction and the polynomial prediction for 0.5 seconds using the training data set. The neural net is specifically trained for the first 0.25 seconds which are flat, after which the performance of the neural network significantly degrades.}
  \label{fig:mse1}
\end{figure}

\begin{figure}[tb]
  \centering
  \includegraphics[width=1\linewidth]{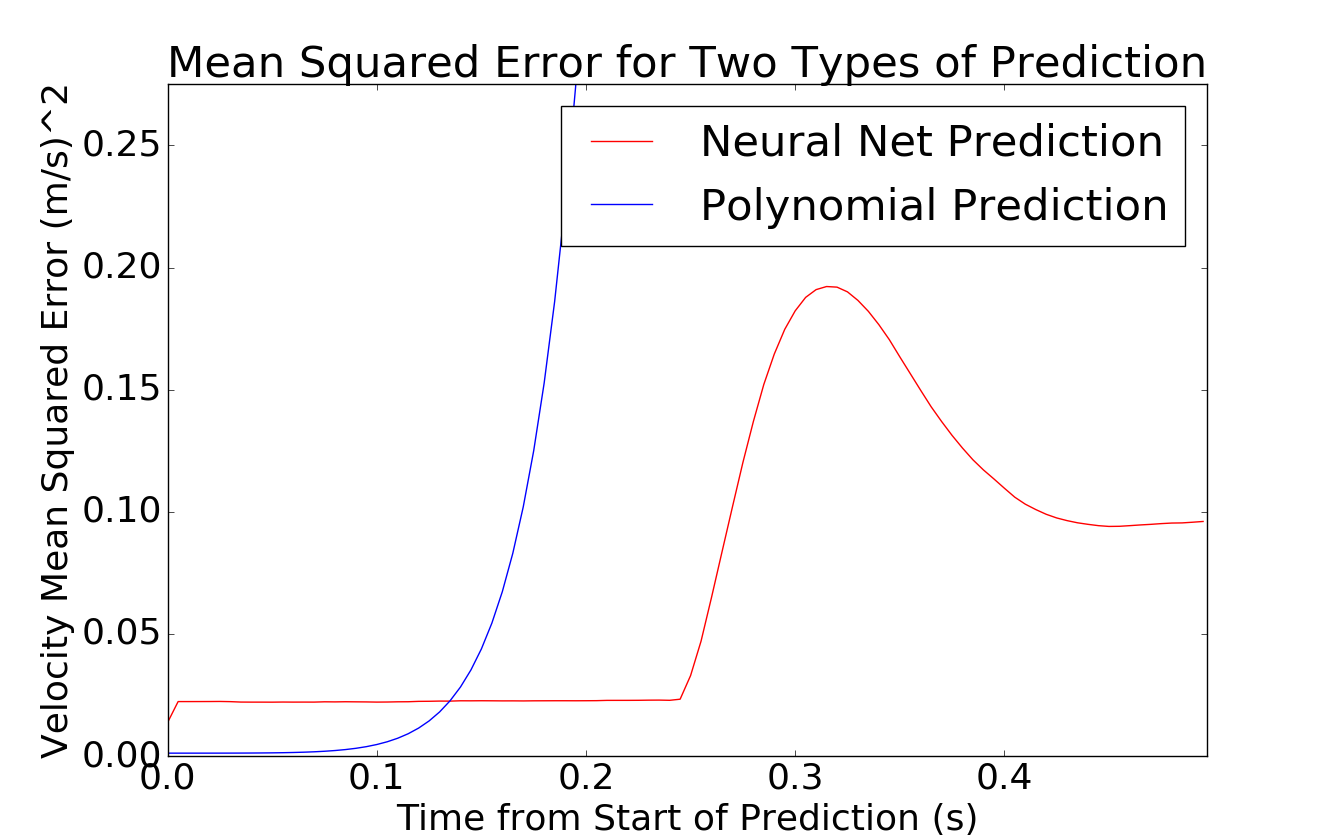}
  \caption{Mean squared error of the neural network prediction and the polynomial prediction for 0.5 seconds using the validation data set. There is very little difference between the prediction on the validation and training data sets.}
  \label{fig:mse2}
\end{figure}

\begin{figure}[tb]
  \centering
  \includegraphics[width=1\linewidth]{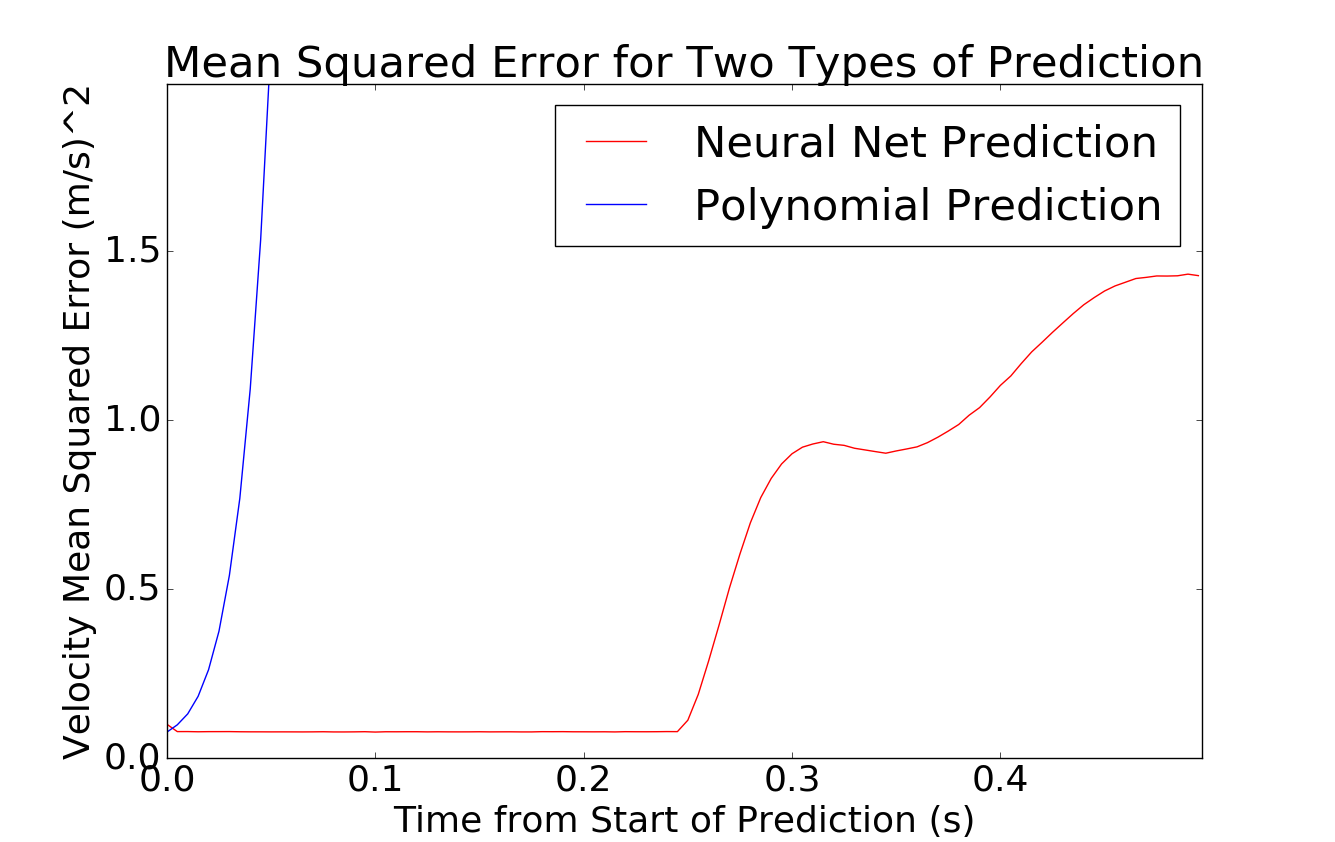}
  \caption{Mean squared error of the neural network prediction and the polynomial prediction for 0.5 seconds using data with added to noise similar to the noise from the encoders on our robot. While both predictions are degraded, the polynomial prediction becomes useless. An 8th order polynomial does not extrapolate well when there is noise.}
  \label{fig:mse3}
\end{figure}

Figure \ref{fig:mse1} shows the MSE of the prediction from the start of the prediction to 0.5 seconds, or 100 time steps, using the data set that was used to train the neural net. Figure \ref{fig:mse2} shows the same thing using the validation set. The similarity of these shows that the neural net does not overfit the data. This is the average of the predictions from every time step. As seen, the polynomial prediction is very good but degrades quickly. The neural net is very accurate for the 0.25 seconds that it was trained for, after which it quickly degrades. Interestingly, the first 50 steps are all predicted with the same accuracy. This is due to the way the neural net is trained. Figure \ref{fig:mse3} shows the MSE with white noise added. The amplitude of the noise is consistent with the noise of the end effector position on our robot. The polynomial prediction degrades very quickly.

\subsection{Estimation with a Robot in the Loop}
\begin{figure}[htb]
  \centering
  \includegraphics[width=1.0\linewidth]{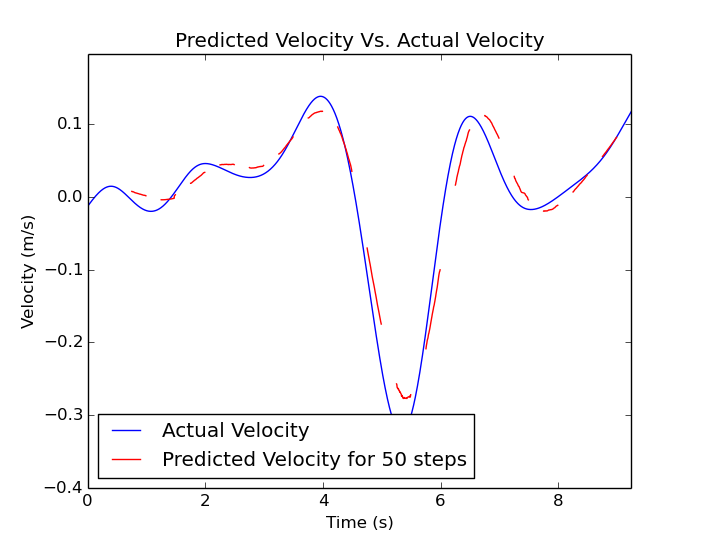}
  \includegraphics[width=1.0\linewidth]{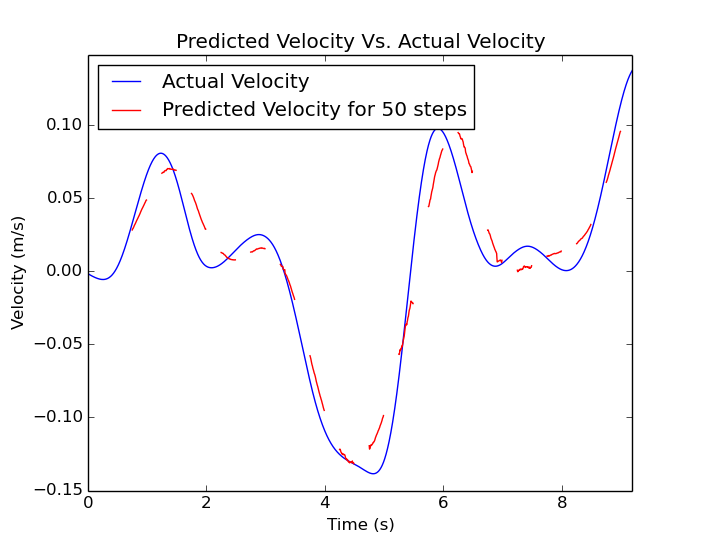}
  \includegraphics[width=1.0\linewidth]{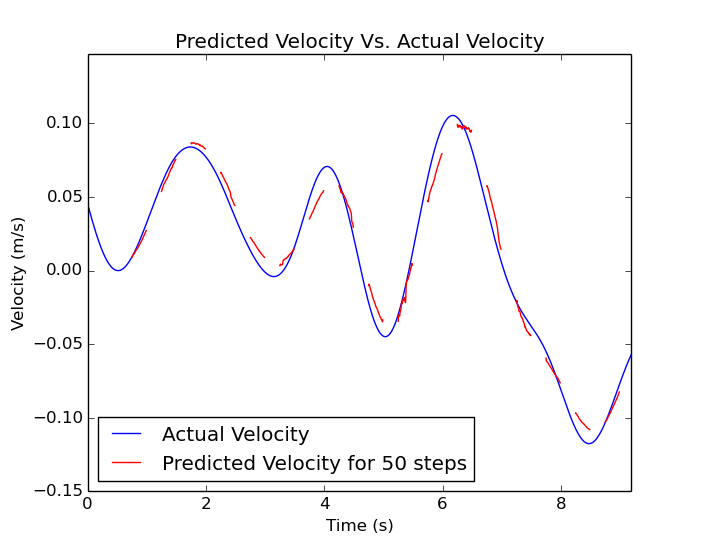}
  \caption{Comparison of velocity prediction to actual future data in
    the x,y, and z directions while a human-robot dyad moves the
    table. Each red line is a separate 50 step prediction using the
    150 steps of real data before it.}
  \label{fig:robot_in_the_loop}
\end{figure}

We ran the estimator while a human-robot dyad moved the table. This was not guaranteed to work because the dyad could not be expected to interact like a human-human dyad. Figure \ref{fig:robot_in_the_loop} shows that the neural net predicted the future motion very well even though the dynamics were different. A robot could use the prediction of human intent to be a better assistant

It is clear that the neural net estimator is able to predict human intent over a short time horizon. There may be additional methods that can improve on this prediction. However, this shows that an accurate and reliable prediction can be made.

\section{Conclusions}
\label{sec:conclusions}
We have shown that human intent can be estimated accurately from
previous motion of the object that is being co-manipulated. We have
also shown that with a mobile robot using a very simple controller in
the loop, the prediction method is still valid. This work lays the
foundation for future controller development which could use the human
intention as a direct input or which could mediate the human intention
and modify it to improve overall performance through haptic or other
means of communication. Ideally, the human and robot would share the
leadership of the task. Advantages for this capability to share
leadership include the following:

\begin{itemize}
\item The robot could be in a position to be a more efficient leader.
\item The robot could better keep away from its own joint limits.
\item The robot could keep the human from violating constraints of the
  task (e.g. the object can not rotate more than 15 degrees of the
  object is person on a stretcher).
\item The robot could have knowledge of the environment not known by the human operator \cite{Evrard2009}. 
\end{itemize}

Interestingly, although even a short prediction into the future should
allow the robot to work better with the human, we have been able to
predict motions over a time horizon that is comparable to human
reaction time. This seems to imply that if used in a closed-loop
control scheme, this estimator would be enough to help a robot perform
more like a teammate and less like a tool. We expect that this
research along with future controller development will allow robots to
work more intuitively and effectively with humans in collaborative
object manipulation tasks.

 



\section*{Acknowledgments}
This research was funded from the Army Research Laboratory Robotics
Collaborative Technology Alliance. We also acknowledge and are
grateful for helpful conversations with Michael Goodrich, Steven
Charles, and Ryan Farrell about our human-human experiments.
 

\bibliographystyle{IEEEtran}
\bibliography{pHRI}
 
\end{document}